\documentclass[conference]{IEEEtran}
\IEEEoverridecommandlockouts
\usepackage[numbers,sort&compress]{natbib}
\usepackage{amsmath,amssymb,amsfonts}
\usepackage{algorithmic}
\usepackage{graphicx}
\usepackage{textcomp}
\usepackage{xcolor}
\usepackage{forest}
\usetikzlibrary{shadows}
\usepackage[colorlinks=false, citebordercolor=green, linkbordercolor=red]{hyperref}
\usepackage[all]{hypcap}

\definecolor{tokenblue}{RGB}{220, 235, 252}
\definecolor{modelgreen}{RGB}{232, 244, 232}
\definecolor{systempurple}{RGB}{238, 226, 240}
\definecolor{rootwhite}{RGB}{255, 255, 255}

\def\BibTeX{{\rm B\kern-.05em{\sc i\kern-.025em b}\kern-.08em
    T\kern-.1667em\lower.7ex\hbox{E}\kern-.125emX}}
\begin{document}

\title{Towards Efficient Large Vision-Language Models: A Comprehensive Survey on Inference Strategies}

\author{\IEEEauthorblockN{ Surendra Pathak}
\IEEEauthorblockA{\textit{Computer Science Department} \\
\textit{George Mason University}\\
Fairfax, VA - 22030, USA \\
spathak8@gmu.edu} \\ 
\and
\IEEEauthorblockN{ Bo Han}
\IEEEauthorblockA{\textit{Computer Science Department} \\
\textit{George Mason University}\\
Fairfax, VA - 22030, USA \\
bohan@gmu.edu}
}

\maketitle

\begin{abstract}
Although Large Vision Language Models (LVLMs) have demonstrated impressive multimodal reasoning capabilities, their scalability and deployment are constrained by massive computational requirements. In particular, the massive amount of visual tokens from high-resolution input data aggravates the situation due to the quadratic complexity of attention mechanisms. To address these issues, the research community has developed several optimization frameworks. This paper presents a comprehensive survey of the current state-of-the-art techniques for accelerating LVLM inference. We introduce a systematic taxonomy that categorizes existing optimization frameworks into four primary dimensions: visual token compression, memory management and serving, efficient architectural design, and advanced decoding strategies. Furthermore, we critically examine the limitations of these current methodologies and identify critical open problems to inspire future research directions in efficient multimodal systems.
\end{abstract}


\section{Introduction}
Large Language Models (LLMs) have emerged as the dominant paradigm for a wide range of natural language processing tasks, demonstrating remarkable capabilities in reasoning, generation, and understanding~\cite{ touvron2023llama, bai2023qwen}. This success has catalyzed a shift toward the development of multimodal systems such as Large Vision Language Models (LVLMs)\cite{liu2023visual, bai2023qwen2, liu2024improved}. By integrating visual encoders with LLM backbones, LVLMs have achieved unprecedented performance on challenging vision-and-language tasks, including visual question answering (VQA), image captioning, and multimodal reasoning. However, the adoption of these models is constrained by their enormous computational demands. For instance, the length of visual tokens increases drastically in LVLMs, introducing a computational burden that scales quadratically with token length. Similarly, the massive influx of visual input sequences from high-resolution images and long videos increases the KV cache footprint. This scenario worsens during the decode phase, where the model repeatedly reads large context from the Key Value (KV) cache, incurring a large latency. Thus, mitigating these memory and computational bottlenecks through efficient inference optimizations has become imperative to enable scalable and real-time application of LVLMs. 

In response to these critical bottlenecks, the research community has explored different mechanisms to utilize compute resources efficiently. These strategies encompass algorithmic enhancements, such as visual token compression, as well as system-level optimizations like KV cache management and LVLM serving. This work presents a comprehensive exploration of the frameworks proposed by the research community to address these challenges. 

The major contribution of this work is to present the current state of visual reasoning inference acceleration paradigms. For that, this review paper systematically outlines effective mechanisms to alleviate computational and memory constraints. In particular, the frameworks are categorized into the following: input context's algorithmic optimizations, system-level enhancements, architectural and structural frameworks, and advanced decoding strategies. Furthermore, these categories are divided into subcategories, as shown in Figure~\ref{fig:taxonomy}, for better readability. In addition to discussing current approaches, this paper points out their limitations. Finally, this work outlines open problems in the visual context inference domain to inspire future research directions.

The remainder of the paper is organized as follows: The preliminaries of the survey are presented in Section~\ref{sec:preliminaries}. The area taxonomy categorizing the optimization frameworks is presented in Section~\ref{sec:area-taxonomy}. After that, the taxonomy-based survey is presented in Section~\ref{sec:taxonomy-survey}. Then, open problems in the domain's current literature are presented Section~\ref{sec:open-problems}. Finally, the concluding remarks are presented in Section~\ref{sec:conclusion}.

\section{Preliminaries}
\label{sec:preliminaries}
\subsection{Large Vision Language Models}
\begin{figure}
    \centering
    \includegraphics[width=0.5\linewidth]{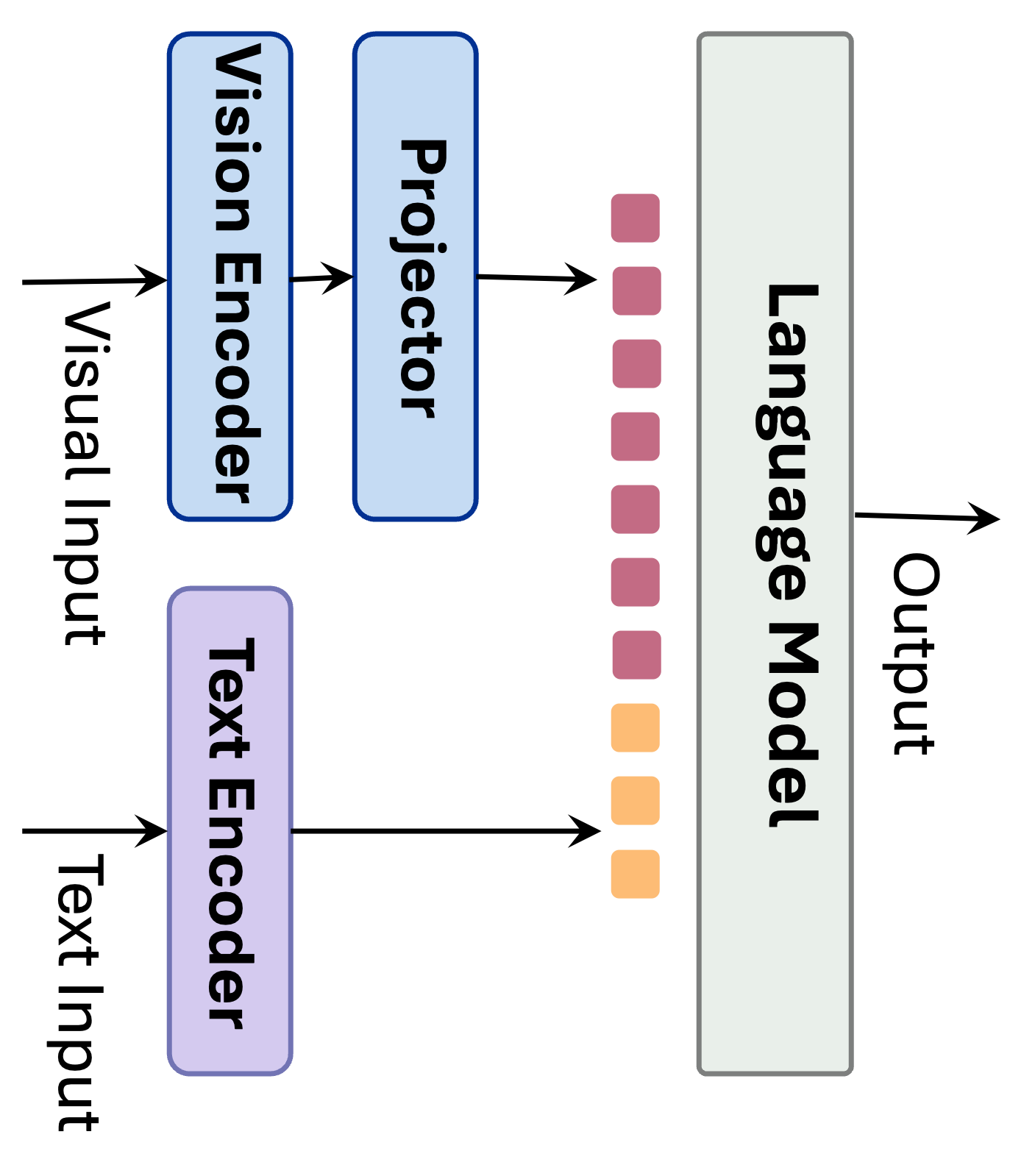}
    \caption{\textbf{Schematic of a standard Large Vision Language Model (LVLM) architecture.} The vision encoder and text encoder convert the visual and text inputs to corresponding tokens. Vision tokens further transform through the projector before being fused with the text tokens. The aggregated tokens are then passed to the backbone language model for autoregressive generation.}
    \label{fig:lvlm-schematic}
\end{figure}

A standard LVLM\cite{liu2023visual, bai2023qwen2, liu2024improved} consists of a vision encoder, a vision-language projector, a text encoder, and a backbone language model. A representative schematic of a standard LVLM model is presented in Figure~\ref{fig:lvlm-schematic}. A vision encoder $g$ compresses an input image $X_v$ into patch features $Z_v$, mathematically defined as $Z_v = g(X_v)$. The most popular vision encoders used in  LVLMs are CLIP~\cite{radford2021learning}, SigLIP~\cite{zhai2023sigmoid}, and InternViT~\cite{chen2024internvl}. Since these encoders constitute a relatively minor portion of a multimodal model's total parameters, the advantages of optimization in this portion of LVLMs are less pronounced. 

A vision-language projector $P$ is a bridge that projects embedded visual patches $Z_v$ into visual embeddings $H_v$ such that $H_v = P(Z_v)$ to align the feature space of the language model. An efficient vision projector must be adaptable to a variable number of input visual tokens and preserve local spatial context with minimal computational overhead. These projected visual representations $H_v$ are concatenated with the encoded textual tokens to form a unified multimodal input sequence, which is passed to the LLM backbone for inference. The LLM backbone is usually a large pre-trained transformer model, which serves as the core reasoning engine of the LVLM. During inference, the model is executed in two stages, namely prefill and decode. Prefill is a compute-bound phase that ingests the entire input sequence in parallel to populate the initial KV cache. Alternatively, the decode phase is a memory-bound, autoregressive process that generates the subsequent output tokens sequentially and is constrained by the latency of repeatedly reading the growing KV cache memory.

\subsection{Self-Attention in Large Language Models}
The backbone LLM~\cite{chiang2023vicuna,touvron2023llama,bai2023qwen,touvron2023llama2,jiang2024mistral} consists of multiple decoder layers that compute causal self-attention values. The input sequence $X$ to each layer consists of system, image, and text tokens given by $X = [x_{\text{sys}}; x_{\text{img}}; x_{\text{txt}}] \in \mathbb{R}^{L \times D}$, where $L$ is the total input length and $D$ is the hidden dimension. Three weight matrices $W_Q, W_K, W_V \in \mathbb{R}^{D \times D}$ are used to compute the query $Q$, key $K$, and value $V$ as:

\[
Q = X W_Q, \quad K = X W_K, \quad V = X W_V
\]

Then, attention value $A$ is computed as:
\[
A(Q, K) = \text{Softmax} \left( \frac{Q K^T + M}{\sqrt{D}} \right),  O = A V.
\]
where $Q \in \mathbb{R}^{L \times D}$ and $K \in \mathbb{R}^{L \times D}$. A lower triangular causal mask $M \in \mathbb{R}^{L \times L}$ is applied to ascertain each token to attend to only the previous tokens and itself. Self-attention in transformer-based LVLMs enables these models to capture a wide range of visual contexts and nuances. However, the quadratic complexity $O(N^2)$ of the attention block to compute attention values during training and inference has become a critical computational bottleneck~\cite{vaswani2017attention}, necessitating the adoption of optimizations to reduce computational demands.

\section{Area Taxonomy}
\label{sec:area-taxonomy}
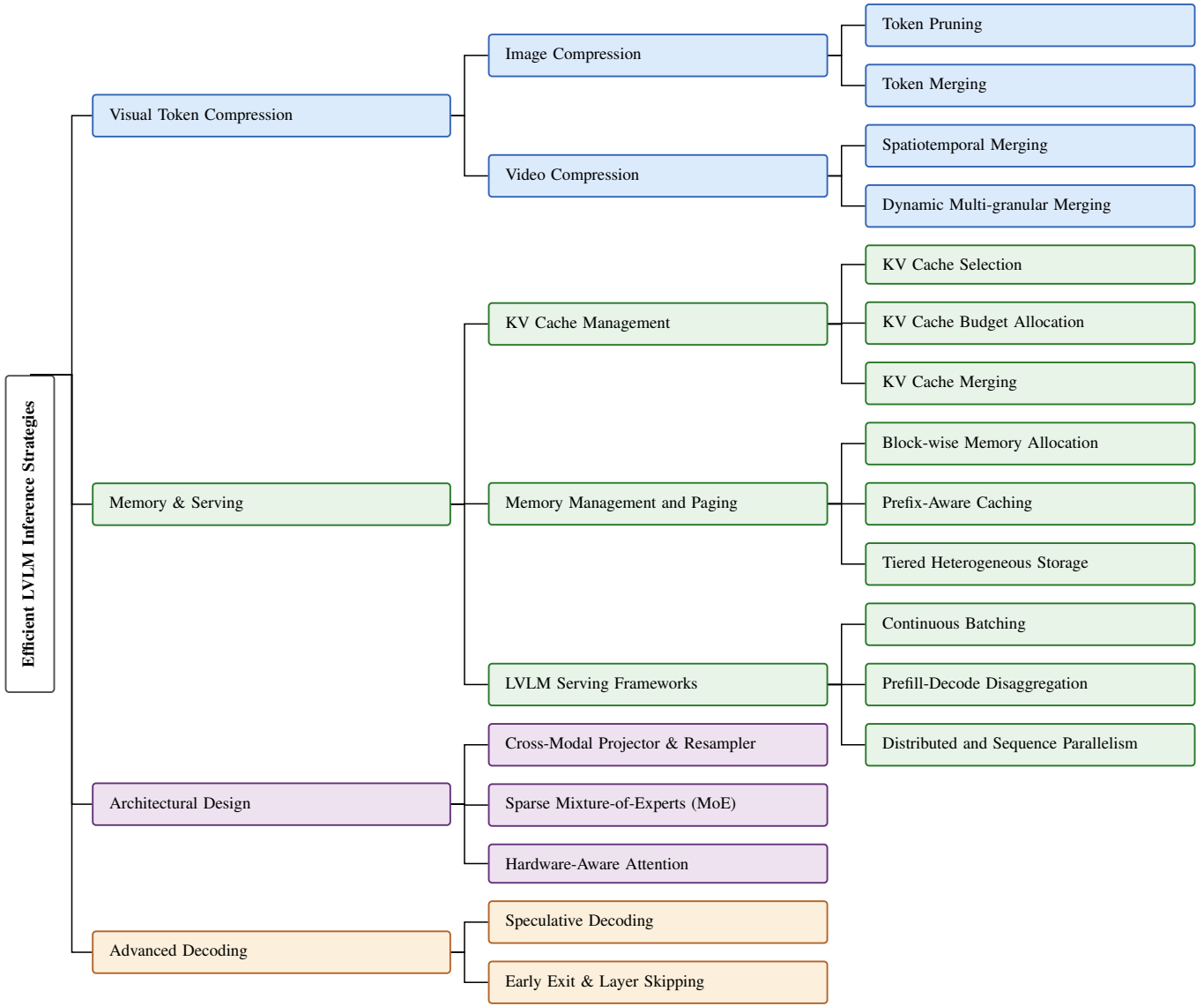
\begin{figure*}[t]
    \centering
    % Background Colors
    \definecolor{color3}{RGB}{220, 235, 252} % Light Blue
    \definecolor{color4}{RGB}{232, 244, 232} % Light Green
    \definecolor{color5}{RGB}{238, 226, 240} % Light Purple
    \definecolor{color6}{RGB}{252, 240, 220} % Light Orange
    
    % Adaptive Border Colors
    \definecolor{border3}{RGB}{50, 100, 180}  % Dark Blue
    \definecolor{border4}{RGB}{40, 120, 40}   % Dark Green
    \definecolor{border5}{RGB}{100, 50, 120}  % Dark Purple
    \definecolor{border6}{RGB}{180, 100, 40}  % Dark Orange
    \definecolor{borderRoot}{RGB}{80, 80, 80} % Dark Gray
    
    \resizebox{\textwidth}{!}{
        \begin{forest}
          for tree={
            font={\normalsize\rmfamily}, % FIXED: Swapped fragile \fontsize for \normalsize
            grow'=0, 
            child anchor=west,
            parent anchor=east,
            anchor=west,
            draw,
            rounded corners=2pt,
            line width=1pt, 
            inner xsep=10pt, 
            inner ysep=8pt,
            l sep=22pt,    
            s sep=10pt, 
            edge path={
              \noexpand\path [draw, line width=0.8pt, \forestoption{edge}] 
              (!u.parent anchor) -- +(8pt,0) |- (.child anchor)\forestoption{edge label}; 
            },
            if level=1{ % FIXED: Level 1 conditional moved safely inside 'for tree'
              edge path={
                \noexpand\path [draw, line width=0.8pt, \forestoption{edge}] 
                (!u.parent anchor) -- +(25pt,0) |- (.child anchor)\forestoption{edge label}; 
              }
            }{}
          }
          % --- Tree Nodes Begin ---
          [Efficient LVLM Inference Strategies, 
            fill=white, 
            draw=borderRoot, 
            rotate=90, 
            anchor=center, 
            minimum height=1.0cm, 
            inner xsep=15pt, 
            inner ysep=10pt,
            font={\normalsize\rmfamily\bfseries} % FIXED
            [Visual Token Compression, fill=color3, draw=border3, text width=6.8cm
                [Image Compression, fill=color3, draw=border3, text width=6.4cm
                    [Token Pruning, fill=color3, draw=border3, text width=6.2cm]
                    [Token Merging, fill=color3, draw=border3, text width=6.2cm]
                ]
                [Video Compression, fill=color3, draw=border3, text width=6.4cm
                    [Spatiotemporal Merging, fill=color3, draw=border3, text width=6.2cm]
                    [Dynamic Multi-granular Merging, fill=color3, draw=border3, text width=6.2cm]
                ]
            ]
            [Memory \& Serving, fill=color4, draw=border4, text width=6.8cm
                [KV Cache Management, fill=color4, draw=border4, text width=6.4cm
                    [KV Cache Selection, fill=color4, draw=border4, text width=6.2cm]
                    [KV Cache Budget Allocation, fill=color4, draw=border4, text width=6.2cm]
                    [KV Cache Merging, fill=color4, draw=border4, text width=6.2cm]
                ]
                [Memory Management and Paging, fill=color4, draw=border4, text width=6.4cm
                    [Block-wise Memory Allocation, fill=color4, draw=border4, text width=6.2cm]
                    [Prefix-Aware Caching, fill=color4, draw=border4, text width=6.2cm]
                    [Tiered Heterogeneous Storage, fill=color4, draw=border4, text width=6.2cm]
                ]
                [LVLM Serving Frameworks, fill=color4, draw=border4, text width=6.4cm
                    [Continuous Batching, fill=color4, draw=border4, text width=6.2cm]
                    [Prefill-Decode Disaggregation, fill=color4, draw=border4, text width=6.2cm]
                    [Distributed and Sequence Parallelism, fill=color4, draw=border4, text width=6.2cm]
                ]
            ]
            [Architectural Design, fill=color5, draw=border5, text width=6.8cm
                [Cross-Modal Projector \& Resampler, fill=color5, draw=border5, text width=6.4cm]
                [Sparse Mixture-of-Experts (MoE), fill=color5, draw=border5, text width=6.4cm]
                [Hardware-Aware Attention, fill=color5, draw=border5, text width=6.4cm]
            ]
            [Advanced Decoding, fill=color6, draw=border6, text width=6.8cm
                [Speculative Decoding, fill=color6, draw=border6, text width=6.4cm]
                [Early Exit \& Layer Skipping, fill=color6, draw=border6, text width=6.4cm]
            ]
          ]
        \end{forest}
    }
    \caption{Taxonomy of Efficient Inference Strategies for Large Vision-Language Models (LVLMs).}
    \label{fig:taxonomy}
\end{figure*}

To systematically present the existing work on the efficient inference of LVLMs, an area taxonomy is presented in Figure~\ref{fig:taxonomy}. The first category, visual token compression, targets the massive influx of input tokens in image and video inferences, addressing the prefill phase's bottleneck. The second category, memory management and serving, comprises system-level optimizations such as KV cache management and disaggregated scheduling frameworks that address the memory-bandwidth constraints of visual workloads. The third category, efficient architectural design, explores the model's structural optimizations, including cross-modal projectors, sparse Mixure-of-Experts (MoE), and hardware-aware attention mechanisms. Finally, advanced decoding strategies consist of frameworks to accelerate the autoregressive generation process with strategies such as multimodal speculative decoding and adaptive computation. In summary, while this classification provides a structure for exploring the current inference optimization strategies, it also provides a framework for identifying the open problems to inspire future research directions.

\section{Taxonomy-Based Survey}
\label{sec:taxonomy-survey}
\subsection{Visual Token Compression}
The primary motivation behind token compression is the inherent feature redundancy observed in visual data. Unlike text, where each word has a distinct role in the sentence, an image may consist of multiple patches that do not have a unique visual feature. For instance, though an image may contain multiple visual patches to represent a sky or a wall, these patches contribute negligible unique semantic value. However, retaining such redundant patches has massive computational implications, especially during the computation of self-attention values. Since standard LVLMs use transformers' self-attention mechanism, which has quadratic computational complexity, retaining such uninformative tokens creates a massive, unnecessary computational bottleneck. The primary objective of visual token compression is to identify and drop/merge such tokens as early as possible to alleviate the computational overhead, while retaining the model's performance.

\subsubsection{Image Token Compression}
Image Token Compression can be classified into image token pruning and image token merging.

\paragraph{Image Token Pruning}
Image token pruning in LVLMs was introduced by the seminal work FastV~\cite{chen2024image}. FastV aggressively discarded visual tokens with low attention scores after the second decoder layer of the backbone LLM. Even though FastV achieves massive computational savings, it occasionally discards visual patches critical for certain fine-grained user prompts due to its task-agnostic mechanism. To address this task-agnostic limitation, subsequent works pivoted towards query-based cross-modal token pruning strategies. For instance, SparseVLM~\cite{zhang2024sparsevlm} evaluates the relevance of visual tokens to the user query, retaining task-critical visual patches. Similarly, TRIM~\cite{song2025less} uses a CLIP metric-based cross-modal pruning strategy to retain tokens based on the user query. 

In addition to the attention-based approaches, alternative strategies that use similarity-based methods to preserve feature diversity have been introduced. For instance, Divprune~\cite{alvar2025divprune} formulates token pruning as a ``Max-Min Diversity Problem'' (MMDP) to select a subset of visual patches that maximizes feature distance. It finds a subset of patches among all possible subsets that has the maximum minimum distance between its elements. Similarly, CDPruner~\cite{zhang2025beyond} uses a determinantal point process (DPP) to maximize the conditional diversity of visual features by maintaining a list-wise diversity based on the pairwise similarity of visual features and user instructions. These approaches discard identical textures, ensuring a diverse, representative set of visual patches is retained for further computation.

Another frontier of research explores alternative techniques to address the limitations of attention-based or diversity-based strategies. For instance, Pyramidrop~\cite{xing2024pyramiddrop} proposed progressively dropping tokens across multiple stages rather than an aggressive, single-stage drop. Similarly, encoder-side methods such as PruMerge~\cite{shang2025llava} and VisionZip~\cite{yang2025visionzip} implement attention-based reduction directly within the visual encoder, eliminating the token sequence even before it reaches the backbone LLM.

\paragraph{Image Token Merging}
While token pruning provides efficiency by permanently eliminating visual information, token merging offers an alternative by consolidating redundant features. The seminal work in this domain was Token Merging (ToMe)~\cite{bolya2022token}, which introduced bipartite soft matching to fuse similar visual points within Vision Transformers. Similar approaches were later adopted for LVLMs in frameworks such as VisionZip~\cite{yang2025visionzip}, which consolidate visual information directly within the visual encoder. Furthermore, hybrid strategies such as PuMer~\cite{cao2023pumer}, ASAP~\cite{pathak2026asap}, and VisPruner~\cite{zhang2025beyond} fuse pruning and merging operations by first pruning uninformative patches and then iteratively consolidating the remaining patches, retaining a condensed visual representation. By aggregating the recurring patterns, such as sky or wall, within an image, these frameworks maintain a feature-dense set for subsequent processing.

\subsubsection{Video Token Compression}
The transition from image to video LVLMs introduced a quadratic increase in computational overhead. For instance, a 10-minute video, though sampled at only 1 FPS, generates a token sequence that rapidly overflows the context window of standard LVLMs. Apart from that, video-based token aggregation poses distinct challenges compared to the prior image-based approaches. For instance, while images contain spatial redundancy, videos further introduce temporal redundancy, necessitating more robust strategies to address this complexity.

Initial techniques extended image token consolidation strategies into the temporal dimension. Then, more dynamic, multi-granular strategies were proposed to address the heterogeneous nature of videos.

\paragraph{Spatiotemporal Merging}
The early frameworks' prominent strategy was to extend existing image-based techniques to the temporal dimension. For instance, Chat-UniVi~\cite{jin2024chat} pooled each frame by applying the K-Nearest Neighbor-based Density Peak Clustering algorithm, i.e., DPC-KNN, to group temporally adjacent frames before merging tokens based on spatial features. Similarly, LLaMA-VID~\cite{li2024llama} consolidated each video frame into two tokens, i.e., one global context token and one local content token, representing temporal and visual cues, respectively. 

While these foundational works successfully prevented Out-of-Memory (OOM) errors, they suffered from several limitations. While relying on single-frame clustering representations misses subtle temporal dynamics, fixed-rate merging algorithms make the approach brittle to the diverse complexity of underlying videos. 

\paragraph{Dynamic, Multi-Granular, and Task-Aware Compression}
The literature quickly recognized the inherent heterogeneity of the underlying videos, such that processing a static frame of a mountain requires less compute than processing a complex action sequence. Subsequent works moved towards this paradigm, presenting dynamic, multi-granular, and task-aware compression frameworks. 

To address the limitations of DPC-KNN, HoliTom~\cite{shao2025holitom} introduced a holistic token condensation approach framing temporal redundancy as a global optimization problem across multiple video frames, rather than a limited local inter-frame issue. Additionally, FastVID~\cite{shen2025fastvid} redefined density-based pruning by dynamically clustering adjacent frames to merge visual patches on-the-fly, while DyCoke~\cite{tao2025dycoke} and Dynamic-VLM~\cite{wang2025dynamic} proposed techniques to flexibly adjust the compression ratio based on the complexity of the video frame sequences. Furthermore, Multi-Granular Spatio-Temporal Token Merging~\cite{hyun2025multi} maintained a multi-level quadtree of video frame patches, which was subsequently used to perform pairwise temporal merging, allowing models to retain patches for complex features while heavily compressing static backgrounds.

One of the prominent shifts in the domain was the cross-modal integration of user prompts and video patches during the consolidation. Such cross-modal approaches were pioneered in LongVU~\cite{shen2025longvu} and FlexSelect~\cite{yunzhuzhang2025flexselect}. Particularly, FlexSelect employed a trained lightweight selector network to mimic the attention weights of a large LVLM, eliminating query-irrelevant visual features before heavy multimodal reasoning occurs. Similarly, LangDC~\cite{wang2025seeing} leveraged a lightweight language model to generate captions from video clips to aid in selecting key visual cues for downstream inference. These cross-modality techniques incorporated textual cues during aggregation to make informed, contextual decisions.

In addition to these approaches, a few other works present new strategies. For instance, FrameFusion~\cite{Fu_2025_ICCV} presented a hybrid prune-merge technique that consolidates identical patches and eliminates unimportant features. Similarly, VQToken~\cite{zhang2025vqtoken} leveraged neural discrete representation learning via vector quantization to map continuous video embeddings into a highly compact feature space, achieving a massive reduction of the visual representation.

\subsection{KV Cache \& Memory Management}
While optimizing the prefill phase via token consolidation addresses the compute bottlenecks, the memory-bound decode phase remains a significant hurdle for real-time autoregressive generation. The implications for LVLM remain critical because high-resolution images and video sequences inject thousands of dense visual tokens into the prefill stage. Consequently, the resulting KV Cache grows linearly when generation progresses, rapidly consuming the GPU High Bandwidth Memory (HBM). This massive memory footprint may lead to the infamous OOM errors, which compromise the model's functionality. Thus, efficient strategies are necessary to address the growing memory footprint. 

\subsubsection{Algorithmic KV Cache Management}
Algorithmic KV cache management can be further categorized into KV cache selection, budget allocation, and merging.

\paragraph{KV Cache Selection}
Algorithmic KV Cache selection strategies that minimize GPU HBM memory consumption by retaining the most critical visual tokens can be categorized into static and dynamic approaches. Static KV Cache selection techniques execute a one-time compression immediately following the LVLMs' prefill phase. For instance, SnapKV~\cite{li2024snapkv} employs an observation window to determine important features, retaining only the critical contextual tokens. Similarly,  L2Compress~\cite{devoto2024simple} establishes a strong correlation between low L2 norms in key embeddings and high attention scores, and compresses the KV Cache based on this observation. 

 Unlike the static approach's single selection during prefill, dynamic selection continuously updates the KV Cache during the decoding phase. StreamingLLM ~\cite{xiao2024efficient}, one of the pioneering works in this category, presented the ``attention sink'' phenomenon, where initial sequence tokens are vital for model stability regardless of their semantic value. Leveraging this observation, StreamingLLM incorporates attention sink positions with recent context for efficient processing. Building on the ``attention sink'' phenomenon, H2O~\cite{zhang2023h2o} dynamically tracked and retained ``Heavy Hitters'' - tokens that accumulate high attention scores over time - while permanently evicting unimportant tokens to enforce a feature-dense KV Cache.

While permanent eviction effectively reduces GPU HBM usage, it irreversibly eliminates information, degrading the model's performance on long contexts. To mitigate this, a subset of dynamic selection approaches avoids permanent eviction by utilizing multi-tier hierarchical storage systems. For instance, InfLLM~\cite{xiao2024infllm} maintains the most critical conversational context in GPU HBM while offloading the less-frequently accessed bulk of KV cache to the host CPU. The offloaded cache is organized with advanced indexing for low-latency retrieval and transfer back to the GPU. For instance, PQCache~\cite{tang2024quest} organizes offloaded cache into block-level representations and uses product quantization codes and centroids to identify important tokens during retrieval. Similarly, SqueezedAttention~\cite{hooper2025squeezed} employs K-means clustering to group semantically related keys on the offloaded cache. During the autoregressive generation loop, query-matching algorithms like SparQ~\cite{ribar2024sparq} dynamically identify and retrieve only the specific top-k relevant KV blocks needed for the current computational step. Such query-matching strategies accelerate the model's access to infinite context without incurring huge latency.

\paragraph{KV Cache Budget Allocation}
Due to the hierarchical architecture of LVLMs, the KV Cache of different layers uniquely contributes to model performance. In that regard, applying a uniform cache compression ratio often becomes suboptimal. To maximize memory utilization while retaining prediction accuracy, KV cache budget allocation strategies distribute memory resources based on the component's importance. These allocation frameworks can be categorized into two levels of granularity: layer-wise and head-wise KV cache budget allocation. 

In contrast to conventional methods that enforce a uniform compression ratio across the entire model, layer-wise budget allocation applies different compression ratios across model layers. PyramidKV ~\cite{cai2025pyramidkv} is a seminal work in this category that implements a pyramid-shaped allocation strategy, i.e., allocating more cache budget in shallow layers and less in the deeper ones. Furthermore, DynamicKV~\cite{zhou-2025-dynamickv} expanded beyond static architectural heuristics to introduce a task-adaptive framework that distributes the memory budget proportionally based on the computed attention-scores of recent and prior tokens across the model layers.  Other frameworks utilize advanced scoring mechanisms to dictate layer budgets. For instance, CAKE~\cite{qin2025cake} frames memory allocation as a ``cake-slicing'' problem, adaptively allocating layer-specific cache sizes based on both the spatial and temporal dynamics of attention.

\paragraph{KV Cache Merging}
To address the growing size of the KV cache as token generation progresses, several merging strategies consolidate similar information. KV Cache aggregation can be done within a layer or across multiple layers.

Intra-layer merging combines representations within individual decoder layers, utilizing either supervised strategies or training-free techniques. Supervised approaches train the model to effectively condense conversation history. For instance, CCM~\cite{kim2024compressed} and LoMA~\cite{wang2024loma} introduce special indicator tokens into the input sequence, e.g., \texttt{⟨COMP⟩} in CCM, acting as checkpoints which are leveraged to compress past key-value pairs in each layer into a compact memory space. Similarly, DMC~\cite{nawrot2024dynamic} employs a learned variable to dynamically decide whether a new KV pair should be added to the cache or merged into an existing KV representation.

Alternatively, training-free methods rely on inherent structural observations to merge tokens without requiring parameter updates. For instance, D2O~\cite{wan2025textdtexto} aggregates the key/value of an evicted token with a retained one using the cosine-similarity-based threshold. Furthermore, CHAI~\cite{agarwal2024chai} 
noted that several heads in multi-head attention that produce correlated attention values can be clustered. Subsequently, CHAI computed attention for only a representative head for each cluster and shared the values across all heads within the group, eventually reducing memory implications.

\subsubsection{Memory Management and Paging}
LVLMs need sophisticated system-level memory management techniques to accommodate the massive influx of visual tokens in modern multimodal architectures. Recent works in this domain generally fall into three key categories: block-wise memory allocation to eliminate fragmentation, prefix-aware caching to reuse shared multimodal contexts, and tiered heterogeneous storage for capacity expansion.

\paragraph{Block-wise Memory
Allocation} The foundational architecture for resolving memory fragmentation during  inference is vLLM~\cite{kwon2023efficient}, which introduced PagedAttention mechanism. Drawing analogy from operating system virtual memory, PagedAttention partitions the KV cache into fixed-size logical blocks and maintains a corresponding block table. The block table maps the continuous sequence tokens to non-contiguous physical blocks in GPU memory. Due to this approach, vLLM can dynamically allocate memory block as per demand rather than pre-allocating maximum possible sequence blocks. Consequently, this strategy eliminates both external fragmentation and the internal memory waste, drastically improving memory utilization and enabling processing more request concurrency without out-of-memory errors.

\paragraph{Prefix-Aware Caching} Beyond minimizing memory fragmentation within individual requests, advanced memory management frameworks address the huge computational redundancy inherent in LVLM inferences via prefix caching. In a multi-turn question answering setup, multiple requests may share identical high-resolution images or system prompts. To exploit the inherent redundancy, prefix caching frameworks organize paged KV blocks into hierarchical structures, enabling dynamic lookup and fetch operations for recurring multimodal data. A leading paradigm in this domain is SGLang~\cite{zheng2024sglang}, which introduced RadixAttention that enables the reuse of precomputed context across multiple generation calls. RadixAttention maintains a Least Recently Used (LRU) cache using a radix tree structure that prioritizes requests based on matched prefix lengths. This is complemented by a strict reference counting mechanism that protects actively used cache entries from premature eviction during continuous batching. Building upon these tree-based data structures, ChunkAttention~\cite{ye2024chunkattention} enables the dynamic detection and sharing of common prompt prefixes during runtime by partitioning monolithic KV cache tensors into smaller, organized chunks. Furthermore, frameworks like BatchLLM~\cite{zheng2024batchllm} enhance system efficiency by co-scheduling requests that share the same KV cache context to maximize cache reuse. Additionally, MemServe~\cite{hu2024memserve} scales these locality-aware prefix policies across distributed environments by pooling CPU and GPU memory across multiple instances to optimize global KV cache sharing.

\paragraph{Tiered Heterogeneous Storage} Even after implementing efficient paging and prefix sharing techniques, ultra-long visual data, such as very long videos, inevitably exhaust the available capacity of the GPU High Bandwidth Memory. To address these storage constraints, tiered memory management extends the effective KV cache capacity by offloading sequence context to the host CPU DRAM and NVMe SSDs, treating them as secondary storage layers. For instance, FlexGen~\cite{sheng2023flexgen} and InfLLM~\cite{xiao2024infllm} implement this framework by continuously monitoring memory states and proactively offloading inactive or historically distant context to the high-capacity secondary storage layers. 
Finally, to alleviate the considerable latency bottlenecks caused by PCIe data transfers, these systems employ asynchronous prefetching techniques that overlap cross-device communication with ongoing GPU computation.

\subsubsection{LVLM Serving}
While memory management and paging  frameworks effectively resolve the storage constraints of massive visual contexts, scheduling the computational resources for these requests presents a unique challenge. To address these new challenges, modern LVLM serving systems introduce sophisticated scheduling mechanisms such as continuous batching, disaggregated execution, and sequence parallelism.

\paragraph{Continuous Batching Strategies} To overcome the constraints of static batching prevalent in prior works, multiple frameworks proposed batching strategies to maximize GPU utilization. For instance, Orca~\cite{yu2022orca} pioneered the continuous batching paradigm where job requests can be dynamically added or removed from a running batch iteratively rather than waiting for the completion of the entire batch. The continuous scheduling strategy was extended by vLLM~\cite{kwon2023efficient}, that integrated iteration-level scheduling with PagedAttention to eliminate memory fragmentation while enabling the concurrent execution of diverse-length job requests within a single batch. Furthermore, FastServe~\cite{wu2023fast} addressed the inherent diverse request lengths by introducing a skip-join Multi-Level Feedback Queue (MLFQ). The MLFQ employs an dynamic scheduling policy that prioritizes shorter requests while demoting heavier long-running requests across multiple priority tiers. This preemptive multi-tier approach minimizes average job completion time while preventing the head-of-line blocking common to static batching pipelines. Building upon these mechanics, Sarathi-Serve introduced chunked prefill, which splits massive compute request sequences into smaller units. It further co-schedules the prefill chunks alongside decoding jobs within a single batch to effectively saturate compute resources.

\paragraph{Prefill-Decode Disaggregation} Although previous batching mechanisms enabled efficient resource utilization, the inherent prefill and decode stages during autoregressive generation pose distinct challenges. Prefill-Decode disaggregation paradigms separate the compute-bound prefill phase and memory-bound decode phase into separate GPU pools, further optimizing GPU usage. Distserve~\cite{zhong2024distserve}, a seminal framework in this domain, disaggregates these phases to meet distinct Service Level Objectives (SLOs), including Time-to-First-Token and Time-per-Output-Tokens, by offloading them to individual computing resources. Furthermore, ShuffleInfer~\cite{hu2025shuffleinfer} adopted this two-level decoupled scheduling strategy, incorporating a component that predicts hardware resource utilization to optimize the allocation of disaggregated workloads. It employs LLM-based prediction to speculate the length of generated tokens of decode requests and schedules them accordingly. Additionally, Infinite-LLM~\cite{lin2024infinite} decouples LLM's attention layers across a pooled GPU memory, facilitating resource sharing to maximize both memory and compute utilization. By analyzing the computational characteristics of dynamic length LLM workloads, Infinite-LLM proposes a mechanism to saturate both compute and memory utilization.

\paragraph{Distributed and Sequence Parallelism} For massive multimodal inputs, the workload may exceed a single GPU's capacity, necessitating a distributed workflow to partition the compute across multiple compute devices. In that regard, distributed and sequence parallelism enable processing ultra-long contexts, unlike the traditional strategies. For instance, DeepSpeed-FastGen~\cite{holmes2024deepspeed} and DeepSpeed-MoE~\cite{rajbhandari2022deepspeed} employ expert parallelism along with Dynamic SplitFuse~\cite{holmes2024deepspeed} to optimize throughput and resource utilization for large-scale deployments. Similarly, RingAttention~\cite{liu2024ringattention} and StripeAttention~\cite{brandon2023striped} distribute self-attention blocks across a ring of GPUs, overlapping cross-device computation and communication, enabling high throughput and low communication overhead. Additionally, an elastic sequence framework was introduced in LoongServe~\cite{wu2024loongserve} that dynamically determines and allocates the GPU count required for a job sequence based on the real-time workload's compute demands. 

\subsection{Efficient Architectural Design}
The architectural components of LVLMs present several opportunities for inference optimization. By redefining the interaction between the vision and language components of the LVLM network, such optimizations can reduce inherent computational overhead, ultimately enhancing resource utilization.

\subsubsection{Cross-Modal Projector \& Resampler Optimization}
Cross-modal projectors and resamplers act as a semantic bottleneck, compressing the massive quantity of visual tokens into a smaller sized latent tokens before passing them to the LLM backbone. Flamingo~\cite{alayrac2022flamingo}, a seminal work in this domain, introduced the ``perceiver resampler'' that maps variable-length visual patches into fixed-sized visual ``queries'' via a cross-attention mechanism. More recently, NVILA~\cite{liu2025nvila} implemented an optimization strategy that processes high-resolution images and videos at full detail initially and then aggressively compresses them into a minimal token set.

\subsubsection{Sparse Mixture-of-Experts (MoE) for LVLMs}
Sparse MoE strategies allow models to increase their parameter count while keeping the token computational cost constant. Specifically, sparse MoE architectures maintain multiple ``experts'' and employ a dynamic routing approach, activating only a subset for any given input sequence rather than using the entire network for every request. In particular, MoE-LLaVA~\cite{lin2026moe}, a seminal framework, established that sparse LVLMs could increase parameter count to model complex visual reasoning capabilities while maintaining the FLOPs count of a smaller model. Additionally, DeepSeek-VL2~\cite{wu2024deepseek} introduced advanced MoE paradigms with fine-grained expert separation and shared experts, enhancing the compute-saving capabilities of MoE architectures. Similarly, the MM1 family of models~\cite{mckinzie2024mm1} validated the effectiveness of these strategies at a massive scale by establishing the comprehensive multimodal scaling laws. By dynamically routing multimodal inputs only to the relevant networks, these MoE-paradigm-based LVLMs support advanced visual reasoning while strictly bounding computational overhead.

\subsubsection{Hardware-Aware Attention Optimization}
The standard self-attention mechanism is highly memory-bound as it repeatedly writes and reads the intermediate attention matrices to and from GPU HBM. FlashAttention~\cite{dao2022flashattention} addressed this bottleneck by introducing an exact attention algorithm that utilizes tiling and forward-pass recomputation to keep the data within on-chip SRAM. Furthermore, FlashAttention-2~\cite{dao2024flashattention} optimized the attention computation by partitioning it between different thread blocks and warps on the GPU, improving parallel occupancy and further reducing the number of non-matmul FLOPs. In addition, FlashAttention-3~\cite{shah2024flashattention} decreases attention computation time on NVIDIA Hopper GPUs by exploiting the compute node's architecture. It does so by overlapping the computation with data movement via wrap-specialization and interleaving block-wise matmul and softmax operations. Overall, the iterations of FlashAttention are the critical hardware-level engine that make processing a massive quantity of high-resolution visual tokens computationally viable.

\subsection{Advanced Decoding Strategies}
After generating the first token in the prefill phase, the LVLM generation transitions to the decode phase, where the subsequent output sequence is generated. At this stage, operations are primarily memory-bound, since the model loads precomputed weights from HBM, often causing a high-latency model pass. To alleviate these limitations, several advanced decoding strategies are introduced.

\subsubsection{Multimodal Speculative Decoding}
Multimodal speculative decoding often employs the ``draft-then-verify'' paradigm, which parallelizes output generation to alleviate the limitations of the memory-bound autoregressive phase. By extending the existing text-only speculative decoding paradigms, a few works have proposed corresponding workflows for multimodal tasks. For instance, Gagrani et al.~\cite{gagrani2024speculative} demonstrated that language-only models can effectively verify multimodal workflows using a 115M language model as a draft for a larger LLaVA 7B LVLM. Additionally, LANTERN~\cite{jang2025lantern} acknowledges the ``token selection ambiguity'' inherent in visual autoregressive models, where models uniformly assign low probabilities to token sequences. LANTERN subsequently mitigates the phenomenon by introducing a relaxed acceptance condition that permits the interchange of semantically similar latent tokens, thereby improving speculative decoding throughput. 

Furthermore, SpecVLM~\cite{ji2025specvlm} extends speculative decoding to video workloads by introducing a training-free paradigm that prunes redundant video patches to minimize the draft model's KV cache footprint. Similarly, ViSpec~\cite{kang2025vispec} employs a tiny vision adaptor to aggregate thousands of redundant visual patches into a compact, representative unit for the draft model, enabling huge throughput gains while maintaining output accuracy. Additionally, DREAM enhances visual speculative decoding by adopting an entropy-adaptive cross-attention fusion mechanism, which injects target model features into the draft model, thereby enhancing representation alignment and achieving massive inference speedups. These frameworks demonstrate that addressing the specialized multimodal context during the drafting process effectively mitigates the memory-bandwidth bottleneck inherent in LVLMs.

\subsubsection{Early Exit \& Layer Skipping}
Adaptive computation strategies, such as early exit and layer skipping, have emerged to address the inherent computational redundancy in LVLMs. Such methods leverage observed token-level complexity variance, where ``easy'' tokens (e.g., punctuation and syntax) can be effectively predicted using only a fraction of the model's layers, whereas ``hard'' tokens require full utilization of the network layers. 

One prominent paradigm of this variance is confidence-based early exits, which introduce internal classifiers to predict the maturity of the token's representation. Such approaches are employed in recent works, such as AdaInfer~\cite{ijcai2025p566}, which utilizes statistical features and classifiers, like SVM and CRF, to terminate computation early for high-confidence predictions. By skipping layers for simple tasks, this strategy can save a large number of computational FLOPs.

\section{Open-Problems}
\label{sec:open-problems}
The research community has presented several solutions to address the growing multimodal input context, memory requirements, and inefficient architectural designs in the LVLM generation scenario. Consequently, these approaches have reduced inference latency and memory footprint, while simultaneously maximizing the model throughput and efficient GPU utilization. However, these works overlook the dynamic shift in the domain towards continuous video streaming, multi-image processing, and real-time agent applications. This section presents such unexplored avenues to define the upcoming research frontiers in vision-language computing efficiency.

While the existing visual compression strategies have successfully mitigated the quadratic complexity of the self-attention block, these approaches are often heuristic-based. Particularly, many of these techniques use attention values, which are susceptible to the decay artifact of rotary positional embedding. Additionally, after the introduction of diversity-based approaches, there is an importance-diversity dilemma on whether to use high-salience important patches or a comprehensive set of diverse patches during complex visual reasoning tasks. Furthermore, the transition to streaming-video-based workloads has presented unique challenges. The live video setup restricts access to future video patches, fundamentally making temporal redundancy elimination a challenging objective. Furthermore, the infinite context length of these inferences becomes a severe memory bottleneck as the KV cache grows linearly with autoregressive generation. Similarly, a visual context eliminated at the current time can eventually be relevant in the future. Since the visual information is already discarded, the model can hallucinate in such circumstances. Designing frameworks that address the importance-diversity dilemma, while maintaining memory-constrained spatial-temporal aggregation in live video streams, remains a critical open challenge for visual token compression.

Although PagedAttention and continuous batching approaches addressed the memory fragmentation for text-based inferences, visual workloads pose distinct challenges. Current KV cache compression techniques often require full attention matrices before deciding which tokens to evict. However, such an approach for high frame rate and long videos results in massive computational waste while computing attention values. An alternative proxy for token salience can be introduced to avoid expensive attention computation. Similarly, existing prefill-decode disaggregation strategies suffer from several limitations for long video or streaming video inferences. While separating the compute-bound prefill and memory-bound decode phases optimizes GPU utilization, transferring the massive visual cache for an inference across the network may incur substantial latency. Due to this, the latency gains from the disaggregation may be diminished. Thus, devising strategies that avoid expensive attention computation and mitigate the overhead of transferring multimodal contexts across disaggregated nodes remains a critical open challenge for scaling LVLM serving systems.

Though the inference efficiency of multimodel LLMs has increased due to the integration of cross-modal projectors, sparse MoE, and hardware-aware attention, several limitations still exist. For instance, the routing algorithm in MoE often routes visual context to a small subset of ``popular'' experts, failing to distribute the tasks across different experts. Consequently, other experts are underutilized, and the model stops functioning like a true mixture of experts. Additionally, even though FlashAttention\cite{dao2022flashattention, dao2024flashattention, shah2024flashattention} optimizes attention for hardware, the underlying mechanism is still $O(N^2)$ complexity, which becomes critical while processing high-resolution images or long videos. While some alternative strategies, such as state-space models~\cite{gu2024mamba} and linear attention~\cite{katharopoulos2020transformers}, mitigate this issue, they suffer from limited expressivity compared to standard attention. These techniques often lose the fine-grained details of the visual content that leads to inadequate performance in complex visual question answering tasks.

\section{Conclusions}
\label{sec:conclusion}
As Large Vision Language Models (LVLMs) handle high-resolution images and continuous video streams, their computational and memory demands become a primary bottleneck for real-world deployment. This paper presents a comprehensive survey of LVLM inference acceleration, organizing existing approaches into four dimensions: visual token compression, memory management and serving, efficient architectural design, and advanced decoding strategies. While current methods have made significant progress in mitigating the quadratic complexity of self-attention and the linear growth of the KV cache, several open challenges exist. In particular, the shift toward continuous video streaming demands sophisticated solutions beyond static heuristics. By identifying these limitations and outlining open problems, this survey offers a roadmap for future research toward scalable, efficient, and robust multimodal systems.

\bibliographystyle{splncs04}
\bibliography{main}

\end{document}